\documentclass[10pt,twocolumn,letterpaper]{article}

\usepackage[pagenumbers]{wacv} 

\usepackage{xcolor}
\usepackage{booktabs}
\usepackage{multirow}
\usepackage{svg}

\newcommand{\qmnew}{\textcolor{black}}
\newcommand{\wqmnew}{\textcolor{black}}

\definecolor{wacvblue}{rgb}{0.21,0.49,0.74}
\usepackage[pagebackref,breaklinks,colorlinks,allcolors=wacvblue]{hyperref}

\def\wacvPaperID{1575} 
\def\confName{WACV}
\def\confYear{2027}

\title{BruNet: A \wqmnew{Cross-Domain} Transfer Framework for Bruise Segmentation}

\author{Qiming Wang\\
Cardiff University\\
United Kingdom\\
{\tt\small WangQ79@cardiff.ac.uk}
\and
Richard J. Motley\\
Cardiff University\\
United Kingdom\\
{\tt\small Richard.Motley@wales.nhs.uk}
\and
Ebube E. Obi\\
Cardiff University\\
United Kingdom\\
{\tt\small ObiE@cardiff.ac.uk}
\and
Xianfang Sun\\
Cardiff University\\
United Kingdom\\
{\tt\small\ SunX2@cardiff.ac.uk}
\and
Paul L. Rosin\\
Cardiff University\\
United Kingdom\\
{\tt\small RosinPL@cardiff.ac.uk}
}

\begin{document}

\maketitle

\begin{abstract}
Segmenting bruises is a challenging task in medical imaging due to limited data and annotations, diffuse boundaries, and highly variable appearance. In this work, we propose \textbf{BruNet}, \wqmnew{a segmentation framework that combines a ViT-based visual encoder (a self-supervised DINOv3 or a pretrained LingBot-Vision backbone) with a SAM-based mask decoder}. BruNet is trained on the HAM10000 skin lesion dataset and evaluated on a separate bruise dataset without additional fine-tuning. Although a small number of prior studies have explored machine learning and computer vision for bruise analysis, existing work has primarily focused on detection, classification, or colour analysis rather than pixel-level localisation. To the best of our knowledge, this is the first study to address automatic bruise segmentation. \qmnew{Our results show that BruNet outperforms CNN-based models, state-of-the-art segmentation models, ChatGPT-4o/5-assisted SAM2 zero-shot baselines, and the medical-oriented MedSAM model, demonstrating strong cross-domain generalisation to bruise segmentation.}
\end{abstract}

\section{Introduction}

Ever since the development of convolutional neural networks (CNNs) and Vision Transformers (ViTs)~\cite{Dosovitskiy2020-fc}, they have replaced traditional statistical and machine learning (ML) methods and become one of the mainstream methods for medical image segmentation tasks. Architectures like U-Net~\cite{Ronneberger2015-vt} and UNETR~\cite{Hatamizadeh2022-hi} demonstrated strong performance on such tasks. However, they often lack generalisability and often require a large amount of annotated high-quality data, which is usually expensive in real-world scenarios, especially when dealing with data-scarce targets such as bruises. Foundation models such as CLIP~\cite{Radford2021-vl} and SAM~\cite{Kirillov2023-mi} offer zero-shot capabilities through semantic alignment and prompt-based segmentation, yet CLIP lacks localisation, and both SAM and CLIP often fail in clinical cases.

Bruise analysis is a representative yet underexplored challenge for medical imaging. Bruises, also referred to as `Contusions' or `Ecchymoses' in clinical contexts, usually exhibit substantial variability in appearance across individuals and over time, with diffuse, poorly standardised boundary definitions that complicate accurate delineation. Moreover, there is a lack of publicly available annotated datasets for bruises, partly because such injuries are often self-managed and rarely clinically documented, particularly in settings such as contact sports where they occur frequently~\cite{Gurau2023-pa}. However, bruising is frequently encountered in clinical practice and may reflect underlying disorders of haemostasis (a process to prevent and stop bleeding), vascular integrity, or systemic disease~\cite{Taylor1981-yx}. In addition, bruises play a critical role in forensic and legal contexts, where accurate documentation and assessment can provide important evidence in cases of physical violence~\cite{MAGGIONI2021101867,Vora488}. With the assistance of computer vision and machine learning, analysing such non-standardised tasks can significantly reduce bias compared to a manual process; the related research remains extremely limited, with existing studies focusing primarily on age classification~\cite{Tirado2021-yz} or detection using bounding boxes~\cite{Aminfar2024-rn}, rather than accurate localisation or segmentation.

In this work, we propose a segmentation framework that integrates a ViT-based backbone, and a SAM decoder to enable cross-domain generalisation. The model is trained on the HAM10000 skin lesion dataset and evaluated on a separate bruise dataset. This method is evaluated against CNN-based baselines, state-of-the-art (SOTA) vision transformer (ViT) segmentation models, and prompt-driven SAM2 variants, \qmnew{including its medical-oriented model MedSAM with MedGemma as prompter~\cite{ma2024segment, sellergren2025medgemma}}. Results show the segmentation robustness under domain shift, highlighting the potential of cross-domain foundation models for dermatological and forensic imaging tasks.

This work also presents the first deep learning framework for bruise segmentation, along with a curated bruise dataset with expert annotations and a novel dual-region protocol for modelling boundary uncertainty.

\qmnew{
Our contributions are summarised as follows:
  \begin{itemize}
      \item We introduce a dual-region annotation protocol for bruise segmentation that separates high-confidence bruise
  core regions from uncertain boundary regions.
      \item We curate and evaluate on an 86-image bruise dataset with expert-reviewed dual-region annotations.
      \item We propose BruNet, a cross-domain segmentation framework trained only on HAM10000 skin lesion masks and
  evaluated without bruise-specific fine-tuning.
      \item We compare BruNet against CNN, ViT, prompt-based SAM2, and medical-oriented baselines under an uncertainty-aware evaluation protocol.
  \end{itemize}
}

\section{Related Works}

\subsection{Vision Transformer and Its Medical Uses}
With the development of deep neural networks (DNNs), they are playing an increasingly vital role in the medical imaging field. Early in this trend, convolutional neural networks like U-Net~\cite{Ronneberger2015-vt} for segmentation, and ResNet~\cite{He_2016_CVPR} for image classification played a dominant role in medical image analysis. However, with the recent development of transformers~\cite{NIPS2017_3f5ee243}, Vision Transformers (ViTs)~\cite{Dosovitskiy2020-fc} with their self-attention mechanisms as backbones, in combination with CNNs~\cite{Hatamizadeh2022-hi}, are starting to outperform traditional CNN models.  Despite their success, most Transformer-based methods still rely heavily on supervised training with large amounts of data, limiting their applicability in data-scarce domains such as rare diseases.

\wqmnew{Self-supervised pretraining addresses this limitation: DINOv3~\cite{simeoni2025dinov3} produces frozen dense representations that transfer to segmentation without labelled data, while LingBot-Vision~\cite{fu2026vision} additionally supervises spatial structure through masked boundary modelling, in which the student reconstructs the semantics and geometry of teacher-identified boundary patches. Both are benchmarked on natural images with well-defined object boundaries, and whether their advantages persist for diffuse, low-contrast targets such as bruises remains untested.}



\subsection{Segment Anything Model}

The Segment Anything Model (SAM)~\cite{Kirillov2023-mi} is a promptable, zero-shot segmentation framework that generalises across domains without task-specific training. SAM supports various inputs as prompts, such as bounding boxes, points, and masks. While SAM demonstrates impressive zero-shot performance on natural images, it underperforms in medical contexts due to differences in features between common objects and disease boundaries. Thus, models like MedSAM~\cite{ma2024segment} were developed by fine-tuning SAM on a large-scale (over 1.5M), annotated medical segmentation dataset. These models improve boundary fidelity and prompt alignment, yet still inherit SAM's underlying assumptions about objectness and edge contrast, which are less reliable in the context of diffuse boundaries. \wqmnew{More recently, SAM 3 extends SAM with open-vocabulary text and exemplar prompts, allowing it to detect and segment objects specified by semantic concepts. This improves its suitability for zero-shot segmentation of previously unseen targets without requiring manually defined spatial prompts~\cite{carion2026sam}}.



\subsection{Computer Vision for Dermatologic Analysis}

Although there is rarely any research done on bruise analysis for computer vision, there are many well-studied works in a similar field. Besides skin lesions, deep learning has been applied to wound classification to support wound assessment and treatment planning~\cite{Anisuzzaman2022-ic}. Skin burn level classification used to be a challenging task for computers, but recent work has achieved high accuracy in identifying burn severity from skin burn images~\cite{SUHA2022100371}.

\qmnew{However, these conditions are not fully equivalent to bruises. Many skin lesions, wounds, and burns manifest as surface-visible abnormalities, often with clearer texture, structural disruption, or boundary cues. Bruises, in contrast, are typically caused by damage to blood vessels beneath the skin, producing subsurface colour changes that may be diffuse, low-contrast, and gradually blended into surrounding tissue. This makes bruise boundaries less well-defined and more difficult to annotate consistently. Therefore, although skin lesion, wound, and burn datasets provide useful visual priors for skin-region analysis, bruise segmentation presents a distinct challenge due to its weaker surface structure and higher boundary ambiguity.}

Despite the advances in many dermatology analysis tasks, the success of deep learning methods is heavily dependent on the availability of large, high-quality annotated datasets. \qmnew{Such datasets are more widely available for skin lesions, wounds, and burns~\cite{lebrat2023syn3dwound,Elsarta2025-al}. For bruises,} data is particularly difficult to obtain. This is largely due to the fact that bruises are often self-managed and under-reported, and in some cases associated with sensitive contexts such as domestic violence, where ethical and privacy constraints limit data sharing and annotation~\cite{Raut2025-tq}.

\subsection{Computer Vision for Bruise Analysis}

Bruises remain a challenging target for machine learning due to their diffuse appearance, inter-subject variability, and sensitivity to lighting, skin tone, as well as skin thickness. Existing work on bruise analysis can be broadly categorised into detection/localisation and feature analysis (e.g. age estimation), while traditional visual assessment by humans remains subjective and prone to bias.

Early studies have explored the use of deep learning for related forensic tasks, including domestic violence classification~\cite{Majumdar_2018_CVPR_Workshops} and injury recognition~\cite{10221832}. More targeted work by Aminfar et al.~\cite{Aminfar2024-rn} adapted lightweight models for bruise detection and examined bias across skin tone classifications~\cite{Desai2025-fo}, highlighting disparities in performance. In parallel, Tirado and Mauricio~\cite{Tirado2021-yz} demonstrated the feasibility of bruise age estimation using CNNs, achieving high accuracy on a large dataset.

Despite these advances, existing approaches focus primarily on detection or classification, with limited attention to precise pixel-level segmentation. This highlights a significant gap in applying modern computer vision methods to bruise localisation.

\section{Methods}

\begin{figure*}
    \centering
    \includegraphics[width=1\linewidth]{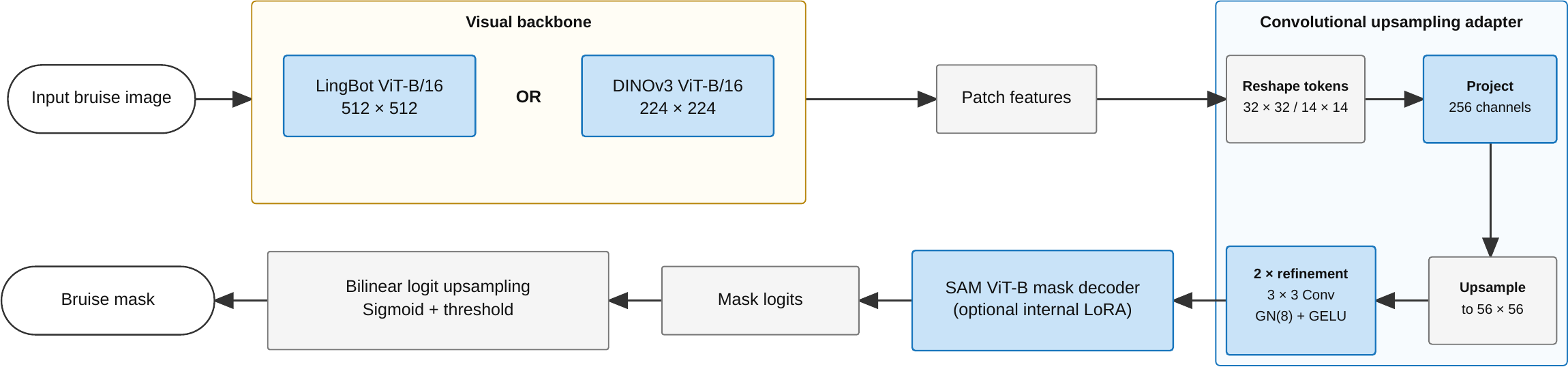}
    \caption{\wqmnew{Overview of the bruise segmentation architecture. LingBot-Vision or DINOv3 encodes the input into a 32 × 32 or 14 × 14 patch-token grid, respectively. The convolutional upsampling adapter reshapes the tokens, projects them to 256 channels, upsamples them to 56 × 56, and applies two 3 × 3 Conv–GN(8)–GELU refinement blocks. The resulting embedding is processed by the SAM ViT-B mask decoder, with optional internal LoRA, before the mask logits are bilinearly upsampled and thresholded to produce the final bruise mask.}}
    \label{fig:fram_examp}
\end{figure*}

In this paper, we propose a structure-aware segmentation framework that integrates a pre-trained vision transformer backbone and the SAM mask decoder to enable promptable, semantically guided segmentation. The model is trained on a pixel-level labelled skin lesion dataset (HAM10000) and evaluated on a separate bruise dataset to assess Zero-Shot Transfer and generalisability under annotation ambiguity, see Fig.~\ref{fig:fram_examp}.

\subsection{Overview}

Our architecture consists of three main components: \wqmnew{(1) a visual backbone — LingBot-Vision ViT-B/16 (masked-boundary pretraining, 512×512 input) or DINOv3 ViT-B/16 (self-supervised, 224×224 input) — as the visual encoder~\cite{fu2026vision, simeoni2025dinov3}, (2) a convolutional upsampling adapter,} and (3) the SAM~\cite{Kirillov2023-mi} (supervised) prompt encoder and mask decoder.

\wqmnew{The visual encoder extracts patch-level features from the input image. Because ViT tokens form a coarse spatial grid, the two backbones emit embeddings at very different resolutions: with a patch size of 16, LingBot-Vision produces a 32 × 32 token grid at its native 512 × 512 input, while DINOv3 produces a 14 × 14 grid at 224 × 224. Reshaping either grid directly into a dense embedding yields a low-resolution feature map that produces masks with visible blocking artefacts, and passes a low-resolution embedding to SAM. We therefore insert a lightweight convolutional upsampling adapter that reshapes the tokens back to their spatial grid, projects them to 256 channels, and upsamples the grid to 56 × 56, followed by two 3 × 3 convolution blocks with group normalisation (8 groups) and GELU activation. The adapter both matches the embedding resolution expected by the SAM mask decoder and recovers fine boundary detail. Crucially, it also acts as a resolution-agnostic interface: both backbones are mapped to a common 56 × 56 × 256 embedding, so the same SAM decoder is used unchanged across variants. The final segmentation mask is produced by SAM from these projected visual tokens.}

\subsection{Training Strategy}

\subsubsection{Training Dataset}

To train the model, we adopt the HAM10000 dataset~\cite{tschandl2018ham10000}, with an expert-annotated binary mask set~\cite{tschandl2020human}, while reserving the 86 annotated bruise images exclusively for evaluation. Due to the scarcity of pixel-level annotated bruise data, we formulate the problem as a Zero-Shot Transfer task, reflecting realistic clinical scenarios where labelled data for rare conditions are limited.

The HAM10000 dataset is one of the largest publicly available dermatology datasets, containing 10,015 high-resolution annotated images, whereas most alternative datasets provide only image-level labels. Although not bruise-specific, skin lesions share key structural characteristics with bruises, including irregular shapes, diffuse boundaries, colour variation, and heterogeneous textures.

By training on diverse lesion morphologies, the model learns boundary-aware and structure-sensitive segmentation priors that are not tied to a specific semantic class. These representations can transfer to bruise segmentation under domain shift, enabling generalisation despite the absence of bruise-specific training data.

\wqmnew{For data augmentation, we noticed that due to the use of dermoscopes while collecting images for HAM10000, the illumination condition changes the appearance of lesions; to allow the model to learn features which are invariant to illumination and contrast, we applied a simple Retinex method~\cite{land1965retinex} as a data augmentation method, specifically, Multi-scale Retinex with Chromaticity Preservation (MSRCP)~\cite{barnard1998investigations}.}


\subsubsection{Training Details}

We trained two variants of our model using \wqmnew{LingBot-Vision ViT-B/16~\cite{fu2026vision} or DINOv3~\cite{simeoni2025dinov3} as visual backbones with all parameters unfrozen, combined with the SAM ViT-B mask decoder adapted via LoRA with rank~$r=8$ and scaling factor~$\alpha=8$, which applies to its attention, MLP, projection, and linear layers.}

\wqmnew{Both variants were optimised using AdamW with a weight decay of~$1\times10^{-4}$ and bfloat16 mixed precision. The DINOv3 model was trained for 150 epochs using a learning rate of~$2\times10^{-5}$ and a batch size of 64. Its input images were resized to~$256\times256$ and centrally cropped to~$224\times224$. The LingBot images were resized to~$576\times576$ and centrally cropped to~$512\times512$. LingBot was trained for 50 epochs, with a learning rate of~$3\times10^{-6}$, and a batch size of 8 with gradient accumulation over eight steps produced an effective batch size of 64.}

The total loss function $\mathcal{L}_{\text{total}}$ combined two components:
\wqmnew{\begin{equation}
\mathcal{L}_{\text{total}} = \mathcal{L}_{\text{Dice}} + \mathcal{L}_{\text{BCE}}
\end{equation}}

The Dice loss encourages overlap between the predicted probability map $P$ and the binary ground truth mask $G$:
\begin{equation}
\mathcal{L}_{\text{Dice}} =
1 - \frac{2 \sum_{i=1}^{N} P_i G_i}{\sum_{i=1}^{N} P_i + \sum_{i=1}^{N} G_i + \epsilon}.
\end{equation}

For BCE, we compute probabilities via a sigmoid, $P=\sigma(z)$, and apply BCE:
\begin{equation}
\mathcal{L}_{\text{BCE}} = - \frac{1}{N} \sum_{i=1}^{N} \left[ G_i \log(P_i) + (1 - G_i) \log(1 - P_i) \right].
\end{equation}

All the above training processes were performed on an NVIDIA GeForce 5070Ti GPU and an Intel i7-12700 CPU with Ubuntu 24.04 LTS via Python 3.11 and PyTorch 2.10.

\section{Experiments}

\subsection{Baselines and Comparisons}

Due to the little to no prior research in this field, we adopted classic CNNs and state-of-the-art segmentation models commonly used in medical imaging and general-purpose vision as baselines for comparison:
\begin{itemize}
    \item \textbf{U-Net (CNN baseline)}: a 2D U-Net baseline~\cite{Ronneberger2015-vt} for binary bruise segmentation, \qmnew{trained also on HAM10000 and evaluated on our bruise dataset}, with four encoder-decoder stages ([64,128,256,512] channels), batch normalisation, ReLU, skip connections, transposed-convolution upsampling, and a single-channel output head. Images were resized to 256 and centre-cropped to 224 with ImageNet normalisation; training used AdamW (lr=1e-4, wd=1e-4, batch size 4, up to 50 epochs, AMP) and a combined Dice + BCE loss. 
    \item \wqmnew{\textbf{OneFormer (ViT baseline)}: a OneFormer model~\cite{jain2023oneformer} with a Swin-L backbone, initialised from the publicly available COCO-pretrained checkpoint and subsequently trained for binary lesion segmentation on HAM10000. Training used $640\times640$ images, AdamW with a learning rate of $1\times10^{-5}$ and weight decay of $1\times10^{-2}$, a batch size of 1, and 30 epochs. The checkpoint with the highest validation IoU was evaluated on our bruise dataset. Binary masks were obtained from the two-class semantic predictions using a threshold of 0.5, without post-processing.}
    \item \wqmnew{\textbf{VLM Prompt-based SAM baselines}: SAM and its variants have been recognised as one of the best segmentation models in many applications as well as the medical field, even without training. To evaluate the effectiveness of our proposed methods,} we evaluated several zero-shot large Vision Language Model prompt-based combinations using GPT-4o, GPT-5, and MedGemma as prompters, paired with either SAM2 or MedSAM as the segmentation model. Specifically, we tested GPT-4o + SAM2, GPT-5 + SAM2, GPT-4o + MedSAM, MedGemma + MedSAM, and MedGemma + SAM2~\cite{sellergren2025medgemma, ma2024segment, hurst2024gpt, singh2025openai}.
    \item \wqmnew{\textbf{SAM3 (zero-shot baseline)}: The latest model from the SAM family~\cite{carion2026sam} was evaluated directly on our bruise dataset using 'skin bruise' as its text prompt, without bruise-specific training or fine-tuning. SAM3 uses open-vocabulary text prompts to detect and segment objects specified by semantic concepts, allowing bruise regions to be segmented without manually defined spatial prompts.}
\end{itemize}

\subsection{Evaluation Setup}

While BruNet was trained solely on skin lesion images, using bruise images was necessary for evaluation. Hence, a ground-truth annotated dataset was created,
\wqmnew{using images searched from various publicly available sources on the internet such as Roboflow and Google Images.}

\qmnew{Due to the diffuse and ambiguous nature of bruise boundaries, defining a single precise ground-truth contour is often clinically unrealistic. The segmentation literature has long recognised this limitation and developed evaluation protocols that operate on multiple ground-truth annotations per image rather than a single reference, most notably the Probabilistic Rand Index (PRI)~\cite{4160946} and its normalised variant, which measure agreement between a prediction and a set of human segmentations through pairwise pixel relations. In medical imaging, methods such as STAPLE~\cite{1309714} complement this view by estimating a probabilistic consensus across expert annotations. In an ideal protocol, multiple medical experts would independently annotate each bruise image, enabling pixel-wise agreement maps to be constructed and PRI-style evaluation to be applied: pixels on which all annotators agreed would form high-confidence bruise regions, while partially overlapping regions would reflect boundary uncertainty.}

\qmnew{However, the data-collection assumptions underlying these protocols are difficult to satisfy in our setting. PRI and related metrics were often adopted on natural-image benchmarks such as BSDS~\cite{937655}, where each image could be annotated at a low cost by several non-expert observers; bruise images, in contrast, require time-consuming annotation by clinically trained experts, making it impractical to obtain a large number of independent annotations per image.}

\begin{figure}[h!]
    \centering
    \setlength{\tabcolsep}{0pt}
    \renewcommand{\arraystretch}{0.6}

    \begin{tabular}{@{}ccc@{}}
        \includegraphics[width=0.315\linewidth]{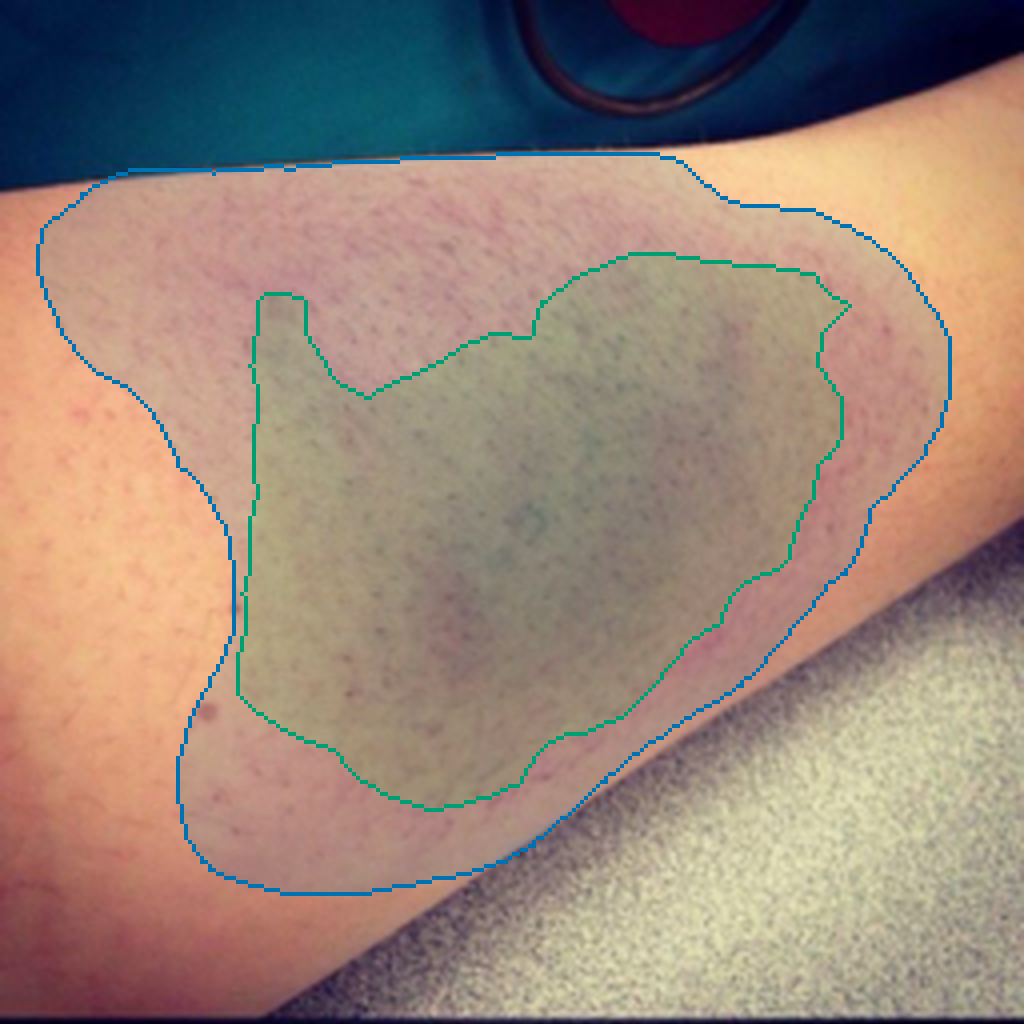} &
        \hspace{0.012\linewidth}
        \includegraphics[width=0.315\linewidth]{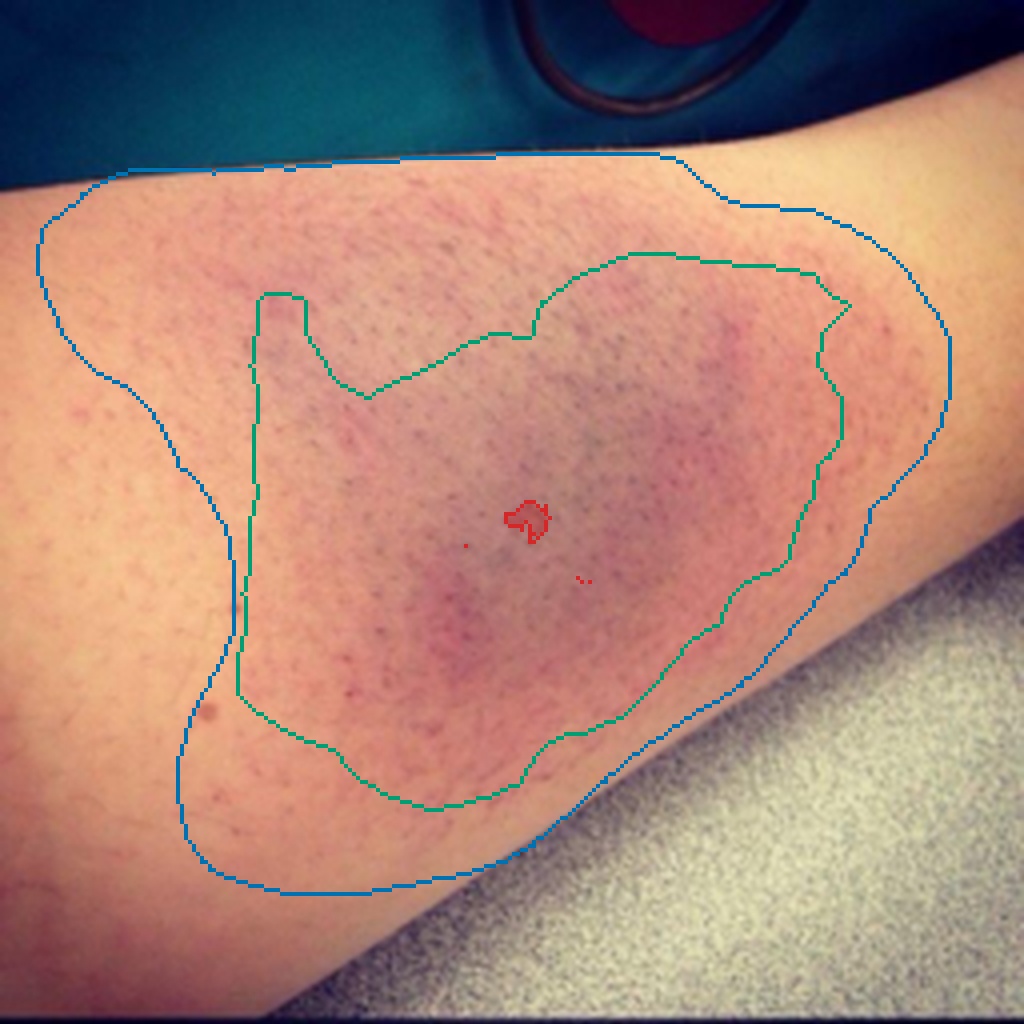} &
        \hspace{0.012\linewidth}
        \includegraphics[width=0.315\linewidth]{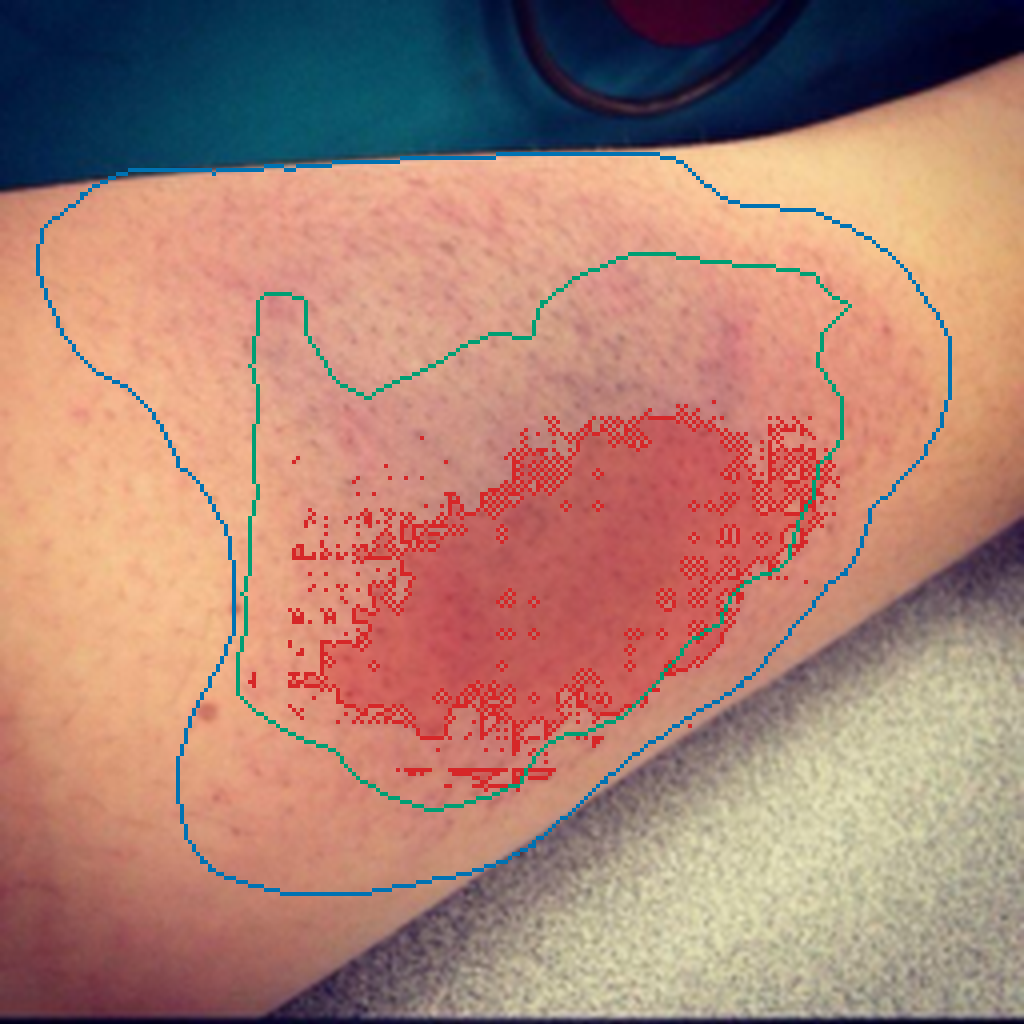} \\

        {\small (a)} & {\small (b)} & {\small (c)} \\

        \includegraphics[width=0.315\linewidth]{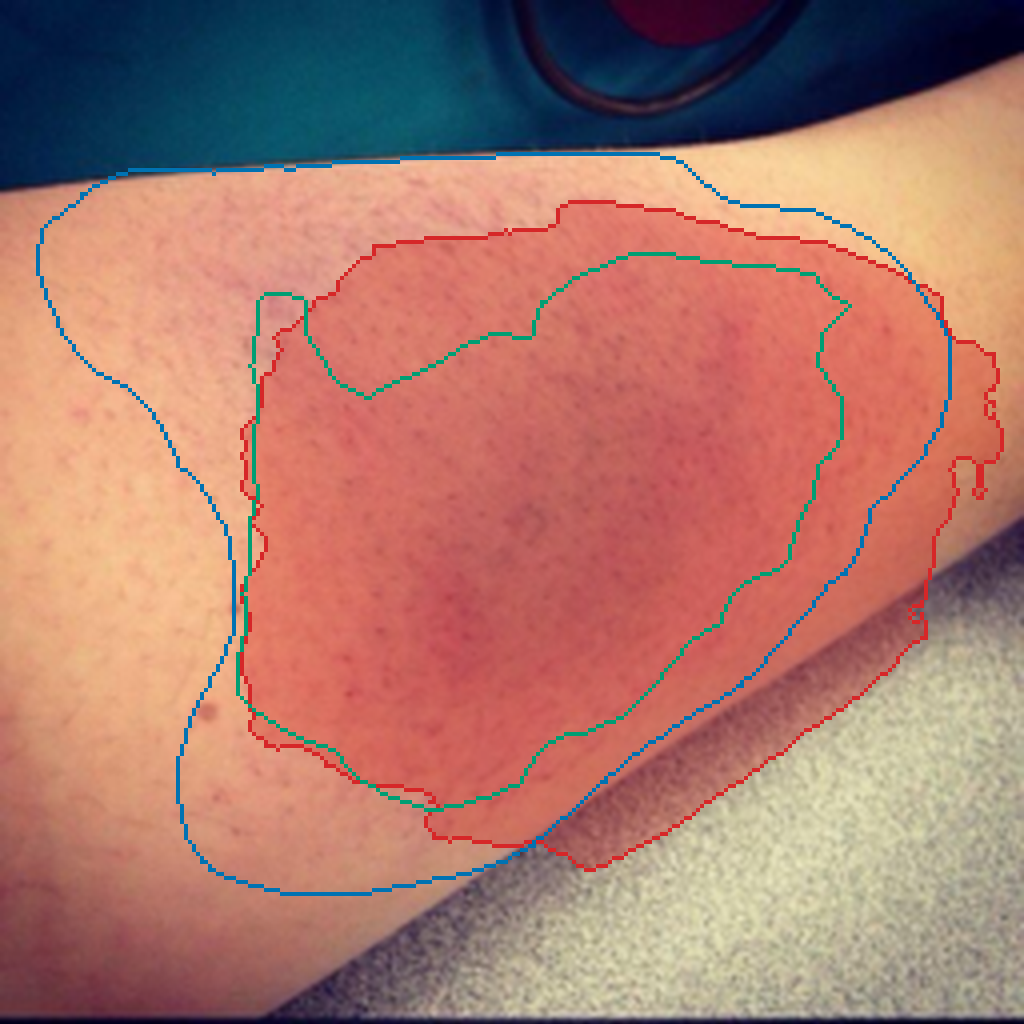} &
        \hspace{0.012\linewidth}
        \includegraphics[width=0.315\linewidth]{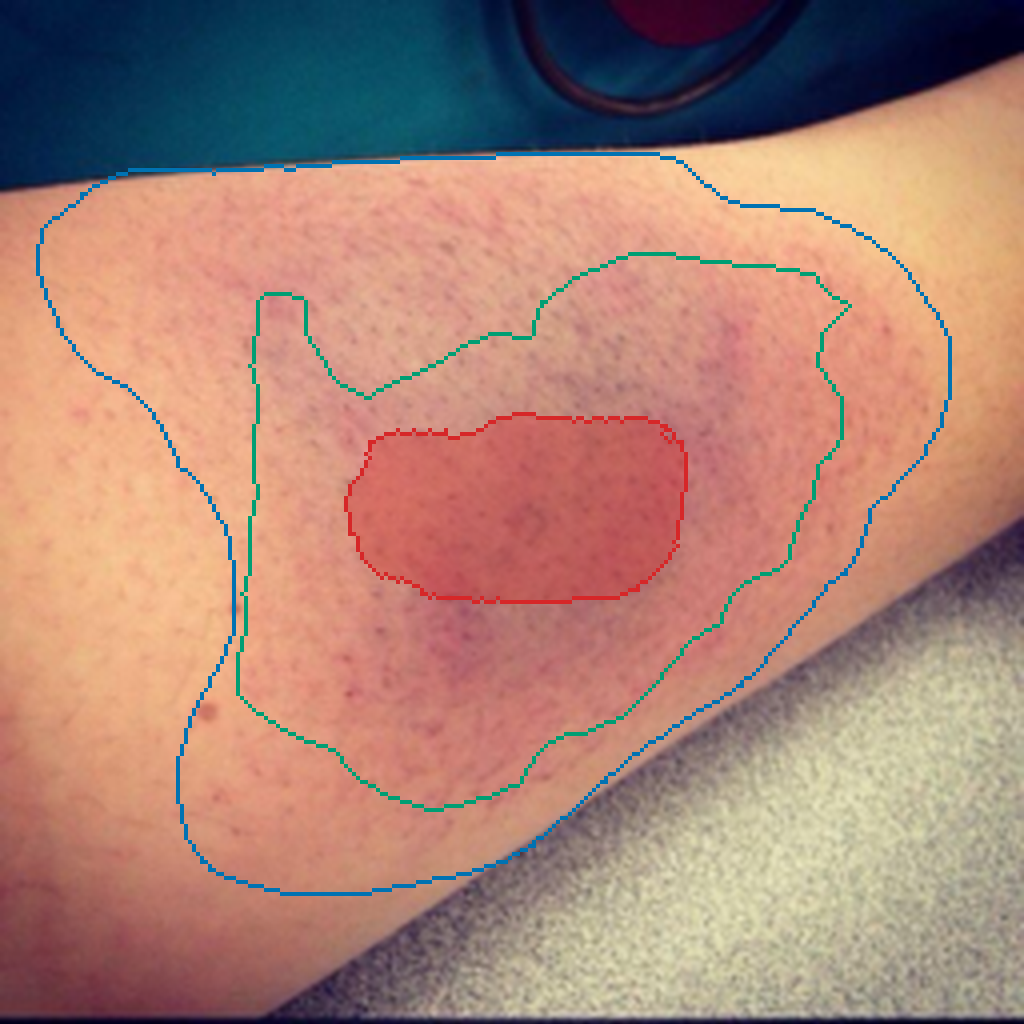} &
        \hspace{0.012\linewidth}
        \includegraphics[width=0.315\linewidth]{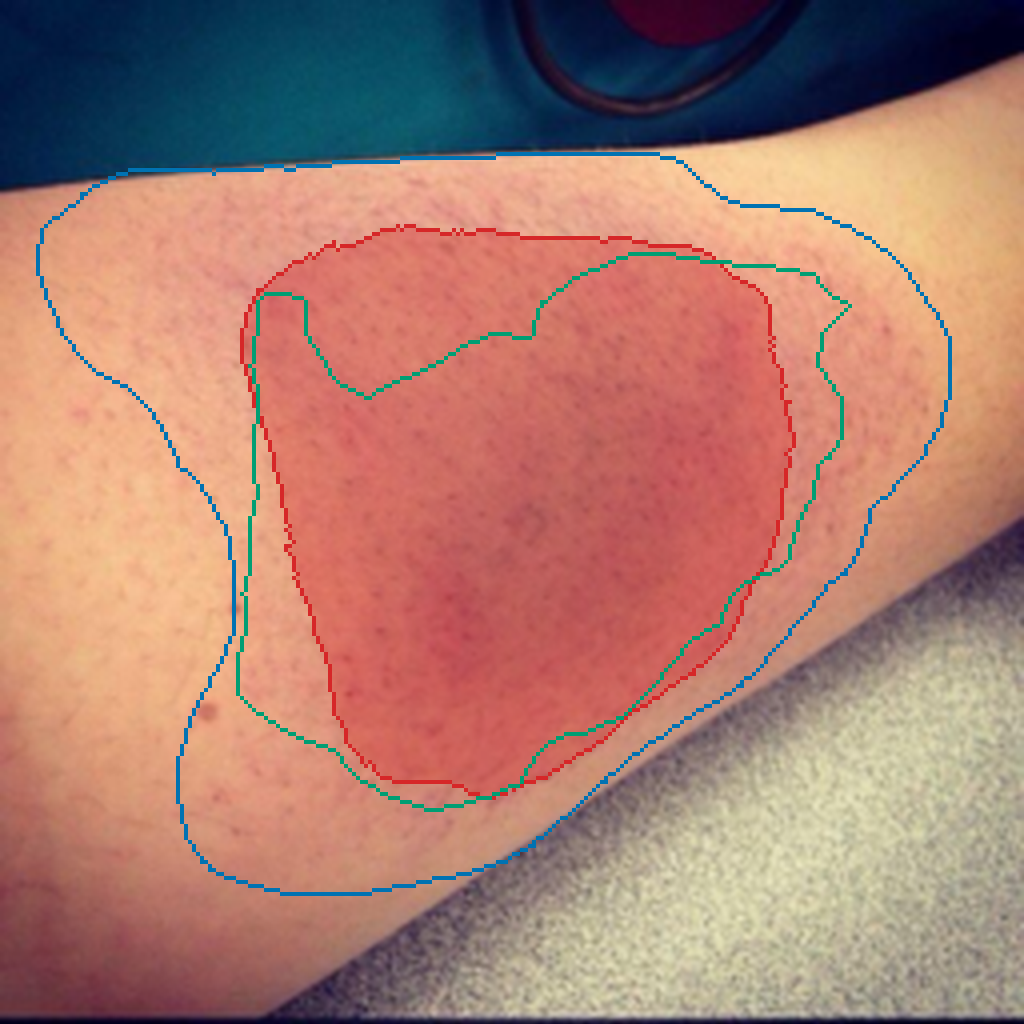} \\

        {\small (d)} & {\small (e)} & {\small (f)} \\

        \includegraphics[width=0.315\linewidth]{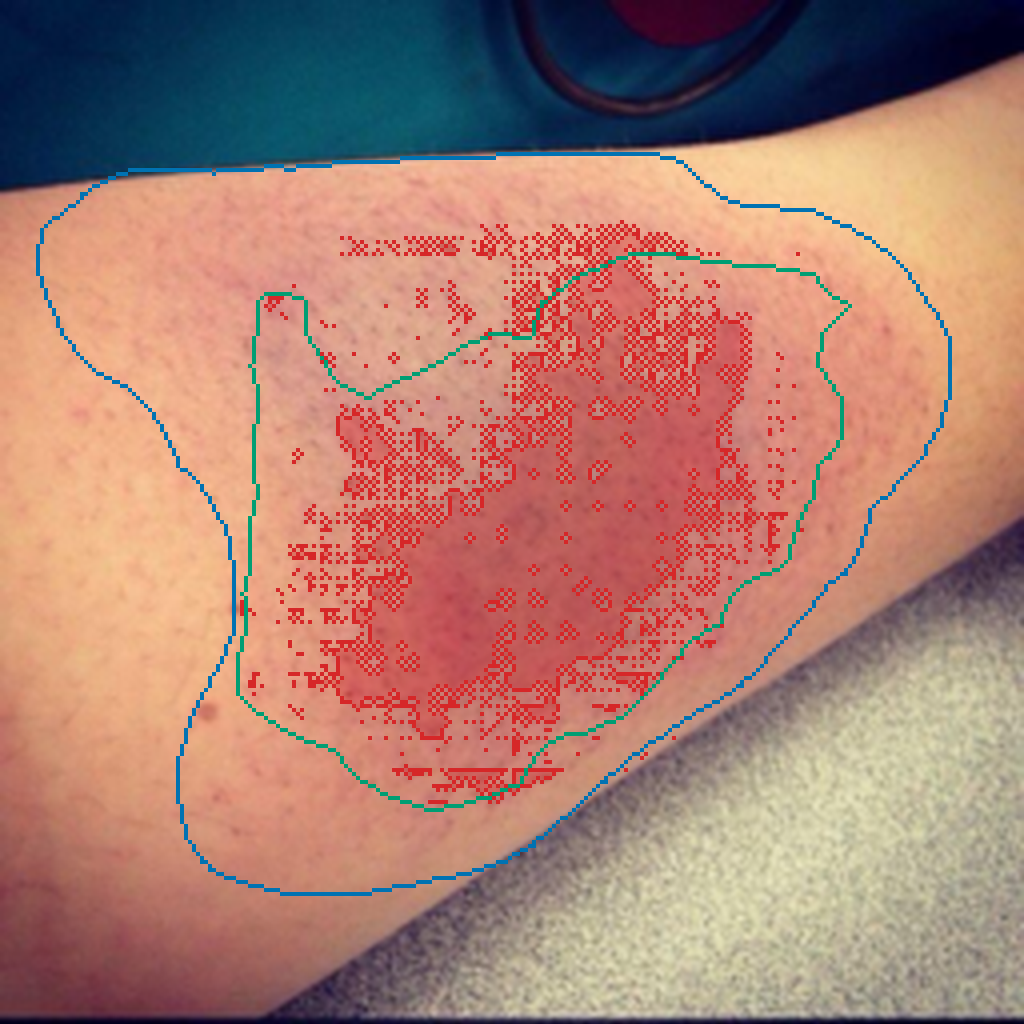} &
        \hspace{0.012\linewidth}
        \includegraphics[width=0.315\linewidth]{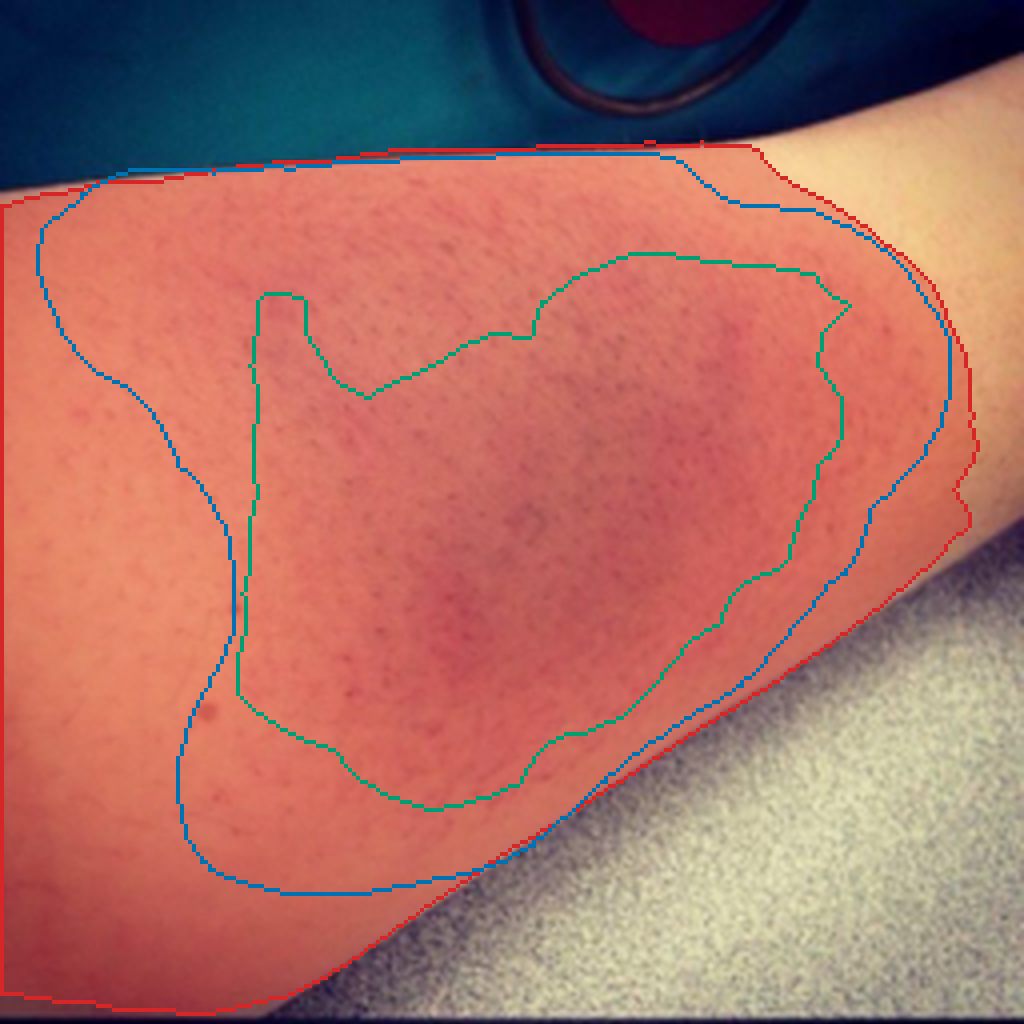} &
        \hspace{0.012\linewidth}
        \includegraphics[width=0.315\linewidth]{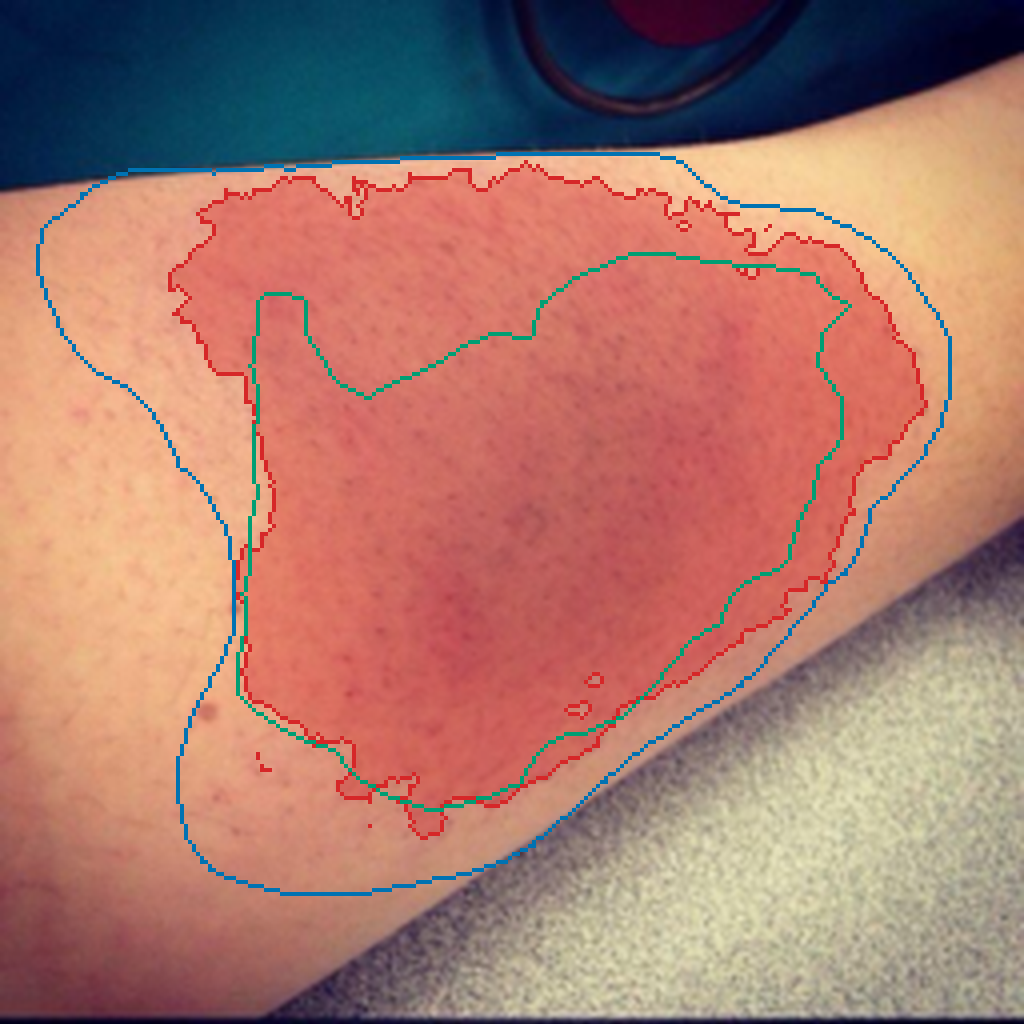} \\

        {\small (g)} & {\small (h)} & {\small (i)} \\
            
        \multicolumn{3}{c}{%
            \begin{tabular}{@{}cc@{}}
                \includegraphics[width=0.315\linewidth]{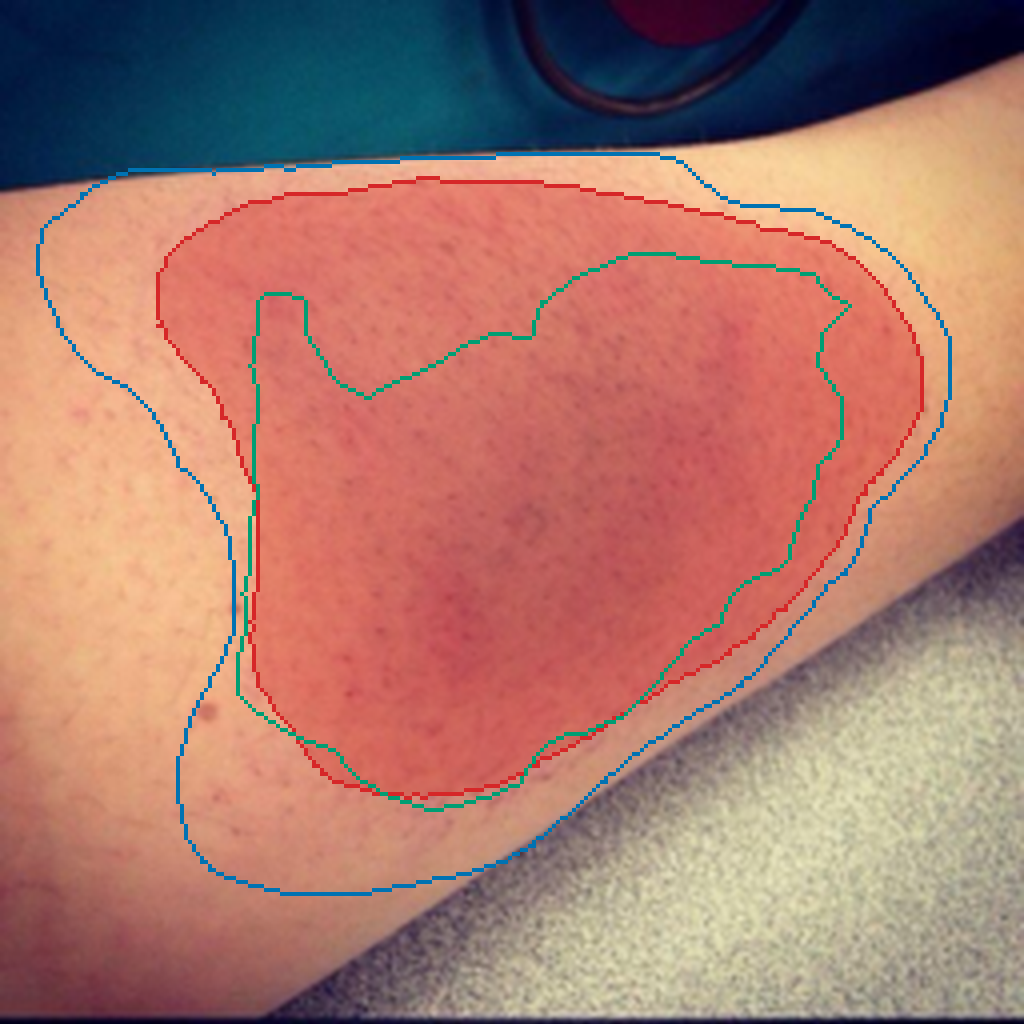} &
                \hspace{0.012\linewidth}
                \includegraphics[width=0.315\linewidth]{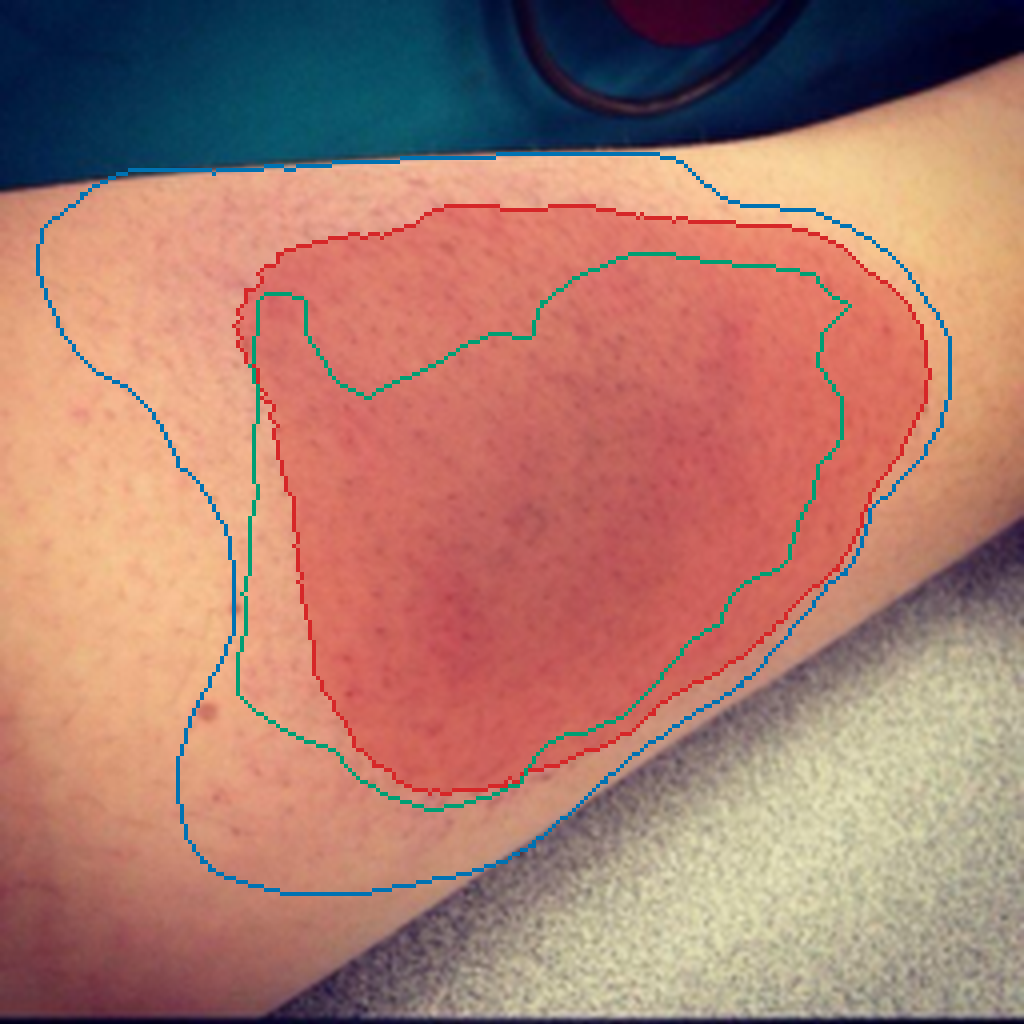} \\
                {\small (j)} & {\small (k)}
            \end{tabular}%
        } \\
        
    \end{tabular}

    \vspace{0.4em}
    \caption{\wqmnew{Visualised comparison of bruise segmentation results. (a) Ground-truth annotation; (b) GPT-4o + SAM2; (c) GPT-5 + SAM2; (d) U-Net; (e) GPT-4o + MedSAM; (f) MedGemma + MedSAM; (g) MedGemma + SAM2; (h) OneFormer; (i) SAM3 zero-shot; (j) BruNet-LingBot (Ours); and (k) BruNet-DINOv3 (Ours). In each prediction panel, the translucent red region represents the predicted bruise mask, while the green and blue contours indicate the inner positive boundary and outer uncertainty boundary, respectively. Pixels between the two ground-truth boundaries were ignored during evaluation and metrics calculation.}}
    \label{fig:mask_compare}
\end{figure}

\qmnew{To approximate uncertainty-aware annotation under these constraints, each image was first annotated by one medical expert, after which a lightweight segmentation model was trained to generate candidate masks at multiple probability thresholds, producing a set of more conservative or more permissive boundary estimates. A second medical expert then reviewed these candidate masks and selected two representative regions: (1) a high-confidence inner region corresponding to the most reliable bruise region, and (2) a broader outer region representing highly possible but uncertain bruise boundaries. This process enabled the construction of a dual-region evaluation protocol that approximates inter-observer uncertainty while remaining feasible under limited annotation resources. An example final dual-region annotation is shown in Fig.~\ref{fig:mask_compare}(a).}

\qmnew{We note that this protocol is a pragmatic surrogate for the multi-annotator standard underlying PRI~\cite{4160946} and STAPLE-style~\cite{1309714} evaluation rather than a substitute for it: the candidate masks reflect model behaviour conditioned on a single expert annotation rather than independent inter-observer variability, and a proper multi-expert annotation study remains an important direction for our future work.}

\subsection{Metrics}

We report two primary standard metrics: 1/ The Dice score, which is widely used in medical image segmentation for clinical relevance, 2/ Intersection-over-Union (IoU), for a more straightforward segmentation performance against other baselines. Both metrics are defined below, where $P$ is the predicted mask, $G$ is the ground truth mask (inside of the inner region), both of which are treated as binary sets:

\begin{equation}
    \text{Dice}(P, G) = \frac{2|P \cap G|}{|P| + |G|}
\end{equation}

\begin{equation}
    \text{IoU}(P, G) = \frac{|P \cap G|}{|P \cup G|}
\end{equation}

\qmnew{For the secondary metrics, we define true positives (TP), false positives (FP), false negatives (FN), and true negatives (TN) at the pixel level. Precision, Recall, and Accuracy are then computed as:}

\begin{equation}
    \text{Precision} = \frac{TP}{TP + FP},
\end{equation}

\begin{equation}
    \text{Recall} = \frac{TP}{TP + FN},
\end{equation}

\begin{equation}
    \text{Accuracy} = \frac{TP + TN}{TP + TN + FP + FN}.
\end{equation}

\qmnew{For Table~\ref{tab:bruise_results}, each metric is calculated independently for each image and then averaged across the 86-image evaluation set. For statistical testing, however, we use per-image Dice and IoU scores, since paired tests require one paired observation per image.}

\qmnew{In the dual-region evaluation, pixels in the high-confidence inner region are treated as positive ground truth, while pixels between the inner and outer regions are excluded from all metric calculations. This prevents uncertain boundary pixels from being classified as either false positives or false negatives.}

\subsection{\wqmnew{Results}}

\begin{table*}[t]
\centering
\caption{\wqmnew{Segmentation performance on the 86-image bruise test set. Each metric was calculated for every image and then averaged across all images. BruNet is compared with U-Net, OneFormer, and zero-shot or prompt-based segmentation baselines.}}
\label{tab:bruise_results}
\begin{tabular}{lccccc}
\hline
\textbf{Model} & \textbf{Precision} & \textbf{Recall} &
\textbf{Dice} & \textbf{IoU} & \textbf{Accuracy} \\
\hline
U-Net                    & 0.4619 & 0.7515 & 0.5198 & 0.4092 & 0.7900 \\
OneFormer                & 0.7063 & 0.8085 & 0.7097 & 0.6352 & 0.9039 \\
GPT-5 + SAM2             & \textbf{0.9467} & 0.2924 & 0.4116 & 0.2922 & 0.8496 \\
GPT-4o + SAM2            & 0.8902 & 0.1497 & 0.2299 & 0.1494 & 0.8178 \\
GPT-4o + MedSAM          & 0.8344 & 0.1943 & 0.2995 & 0.1916 & 0.8236 \\
MedGemma + MedSAM        & 0.6025 & 0.5771 & 0.5157 & 0.3911 & 0.8472 \\
MedGemma + SAM2          & 0.6575 & 0.2846 & 0.3491 & 0.2632 & 0.8479 \\
SAM3 (zero-shot)         & 0.9372 & 0.5888 & 0.7053 & 0.5875 & 0.9208 \\
\hline
BruNet-DINOv3 (Ours)     & 0.8691 & \textbf{0.8491} & 0.8386 & 0.7592 & 0.9441 \\
BruNet-LingBot (Ours)    & 0.9464 & 0.8227 & \textbf{0.8674} &
\textbf{0.7897} & \textbf{0.9556} \\
\hline
\end{tabular}
\end{table*}

Table~\ref{tab:bruise_results} reports the mean results across the 86 test images. Both BruNet variants achieved higher Dice and IoU scores than every evaluated baseline. BruNet-LingBot obtained the highest Dice (0.8674), IoU (0.7897), and accuracy (0.9556), whereas BruNet-DINOv3 achieved the highest recall (0.8491) and the second-highest Dice (0.8386) and IoU (0.7592). Moreover, neither variant was trained on bruise images, yet both substantially outperformed the evaluated baselines. These results demonstrate the effectiveness of adapting pre-trained visual representations and the SAM mask decoder for cross-domain transfer from skin lesion segmentation to bruise segmentation.

Among the conventional segmentation baselines, OneFormer achieved the strongest Dice (0.7097) and IoU (0.6352), supported by a high recall of 0.8085. However, its lower precision of 0.7063 indicates a greater tendency to include non-bruise regions. This behaviour is visible in Fig.~\ref{fig:mask_compare}(h), where the prediction extends beyond the annotated outer boundary. U-Net achieved a recall of 0.7515 but a precision of only 0.4619, resulting in a Dice score of 0.5198 and an IoU of 0.4092. Its prediction in Fig.~\ref{fig:mask_compare}(d) similarly extends outside the outer boundary, illustrating its tendency towards over-segmentation.

The LLM-prompted SAM2 and MedSAM pipelines generally produced conservative or incomplete masks. GPT-5 + SAM2 achieved the highest precision among all evaluated models (0.9467), but its low recall (0.2924) limited its Dice and IoU to 0.4116 and 0.2922, respectively. GPT-4o + SAM2 was even more conservative, with a recall of 0.1497, a Dice score of 0.2299, and an IoU of 0.1494. As illustrated in Fig.~\ref{fig:mask_compare}(b) and (c), these predictions capture only limited portions of the bruise. Among the five LLM-prompted pipelines, MedGemma + MedSAM performed best, obtaining a Dice score of 0.5157 and an IoU of 0.3911. Nevertheless, its performance remained below OneFormer, SAM3, and both BruNet variants. These results suggest that bounding-box localisation followed by SAM-style segmentation often fails to recover the complete extent of diffuse bruises. However, the current evaluation does not isolate whether this limitation originates from bounding-box localisation, prompting, or mask generation.

Native text-prompted SAM3 achieved high precision (0.9372) and the highest accuracy among the baselines (0.9208). Its lower recall of 0.5888 reduced its Dice and IoU to 0.7053 and 0.5875, respectively. Its Dice score was close to that of OneFormer, although its IoU was lower. SAM3 produced a strong segmentation for the example in Fig.~\ref{fig:mask_compare}(i), but it returned no retained bruise instance for five of the 86 test images. Therefore, the selected visualisation does not represent its performance on every image.

In comparison, the BruNet predictions in Fig.~\ref{fig:mask_compare}(j) and (k) form spatially coherent bruise regions that closely follow the annotated boundaries. BruNet-LingBot provides the strongest overall overlap and precision--recall balance, while BruNet-DINOv3 achieves slightly greater bruise coverage, as reflected by its higher recall. The strong performance of both variants supports the use of pretrained visual representations adapted with the SAM mask decoder for segmenting bruises characterised by weak contrast, diffuse colour changes, and uncertain boundaries. \wqmnew{Furthermore, our best model, BruNet-LingBot, requires only \raisebox{-0.5ex}{\textasciitilde}400 MiB of VRAM and takes at most 30\,ms per image on our experimental setup, while one of the better-performing baselines, SAM 3, requires over 4000 MiB of VRAM and more than 230\,ms per image. This demonstrates BruNet-LingBot's substantially higher computational efficiency and its greater suitability for real-world applications.}

\subsubsection{\wqmnew{Statistical Analysis}}

To assess whether the observed performance differences were statistically reliable, we applied two-sided paired Wilcoxon signed-rank tests~\cite{conover1999practical} to the per-image Dice and IoU scores from the 86-image evaluation set. Pairing was used because every model was evaluated on the same images. Table~\ref{tab:bruise_results} reports the arithmetic mean of these per-image scores, whereas the statistical tests operate on matched image-level differences between models.

For each model, we estimated a 95\% confidence interval for the mean Dice and IoU using 10,000 nonparametric percentile-bootstrap resamples of the images. BruNet-DINOv3 was retained as the reference model for the paired comparisons. For each comparison, $\Delta_{\mathrm{med}}$ denotes the median of the 86 paired differences, calculated as the BruNet-DINOv3 score minus the comparison-model score. Its 95\% confidence interval was obtained by paired bootstrap resampling. Zero differences were excluded from the signed-rank calculation using the Wilcox convention. Holm--Bonferroni correction~\cite{holm1979simple} was applied separately to the nine Dice comparisons and the nine IoU comparisons. Statistical significance was assessed at $\alpha=0.05$. A fixed random seed of 20260824 was used for reproducibility.

\begin{table*}[t]
\centering
\caption{Bootstrap 95\% confidence intervals for the mean per-image Dice and IoU scores across the 86-image evaluation set.}
\label{tab:bootstrap_ci}
\begin{tabular}{lcc}
\hline
\textbf{Model} & \textbf{Dice mean (95\% CI)} & \textbf{IoU mean (95\% CI)} \\
\hline
U-Net & 0.5198 (0.4572--0.5841) & 0.4092 (0.3500--0.4719) \\
OneFormer & 0.7097 (0.6398--0.7776) & 0.6352 (0.5646--0.7058) \\
GPT-5 + SAM2 & 0.4116 (0.3601--0.4657) & 0.2922 (0.2479--0.3396) \\
GPT-4o + SAM2 & 0.2299 (0.1856--0.2755) & 0.1494 (0.1150--0.1860) \\
GPT-4o + MedSAM & 0.2995 (0.2606--0.3390) & 0.1916 (0.1612--0.2234) \\
MedGemma + MedSAM & 0.5157 (0.4583--0.5718) & 0.3911 (0.3403--0.4428) \\
MedGemma + SAM2 & 0.3491 (0.2812--0.4174) & 0.2632 (0.2071--0.3204) \\
SAM3 (zero-shot) & 0.7053 (0.6527--0.7539) & 0.5875 (0.5374--0.6369) \\
\hline
BruNet-DINOv3 & 0.8386 (0.7962--0.8766) & 0.7592 (0.7100--0.8058) \\
BruNet-LingBot & 0.8674 (0.8348--0.8963) & 0.7897 (0.7488--0.8278) \\
\hline
\end{tabular}
\end{table*}

\begin{table*}[t]
\centering
\caption{Paired comparisons between BruNet-DINOv3 and each other model on the 86-image evaluation set. $\Delta_{\mathrm{med}}$ is the median paired difference (BruNet-DINOv3 minus the comparison model), reported with a paired-bootstrap 95\% confidence interval. The two-sided Wilcoxon signed-rank $p$-values were Holm--Bonferroni corrected separately across the nine comparisons for each metric. $^{*}$ denotes significance at $\alpha=0.05$ after correction.}
\label{tab:stat_tests}

\begin{minipage}[t]{0.49\textwidth}
\centering
\textbf{(a) Dice}

\vspace{0.3em}

\resizebox{\linewidth}{!}{%
\begin{tabular}{lcc}
\hline
\textbf{Comparison model} &
\textbf{$\Delta_{\mathrm{med}}$ (95\% CI)} &
\textbf{$p_{\mathrm{adj}}$} \\
\hline
U-Net & +0.2676 (+0.2093, +0.3198) & $<0.001^{*}$ \\
OneFormer & +0.0125 (-0.0055, +0.0627) & 0.0044$^{*}$ \\
GPT-5 + SAM2 & +0.4564 (+0.3504, +0.4957) & $<0.001^{*}$ \\
GPT-4o + SAM2 & +0.6682 (+0.6037, +0.7487) & $<0.001^{*}$ \\
GPT-4o + MedSAM & +0.5774 (+0.5170, +0.6225) & $<0.001^{*}$ \\
MedGemma + MedSAM & +0.2705 (+0.2373, +0.3575) & $<0.001^{*}$ \\
MedGemma + SAM2 & +0.5120 (+0.3586, +0.6213) & $<0.001^{*}$ \\
SAM3 (zero-shot) & +0.1128 (+0.0975, +0.1327) & $<0.001^{*}$ \\
BruNet-LingBot & -0.0010 (-0.0132, +0.0126) & 0.4096 \\
\hline
\end{tabular}%
}
\end{minipage}
\hfill
\begin{minipage}[t]{0.49\textwidth}
\centering
\textbf{(b) IoU}

\vspace{0.3em}

\resizebox{\linewidth}{!}{%
\begin{tabular}{lcc}
\hline
\textbf{Comparison model} &
\textbf{$\Delta_{\mathrm{med}}$ (95\% CI)} &
\textbf{$p_{\mathrm{adj}}$} \\
\hline
U-Net & +0.3167 (+0.2561, +0.3733) & $<0.001^{*}$ \\
OneFormer & +0.0220 (-0.0110, +0.0913) & 0.0059$^{*}$ \\
GPT-5 + SAM2 & +0.4847 (+0.4049, +0.5689) & $<0.001^{*}$ \\
GPT-4o + SAM2 & +0.6738 (+0.5851, +0.7704) & $<0.001^{*}$ \\
GPT-4o + MedSAM & +0.6425 (+0.5433, +0.6945) & $<0.001^{*}$ \\
MedGemma + MedSAM & +0.3792 (+0.3038, +0.4493) & $<0.001^{*}$ \\
MedGemma + SAM2 & +0.5011 (+0.4220, +0.6115) & $<0.001^{*}$ \\
SAM3 (zero-shot) & +0.1719 (+0.1441, +0.1966) & $<0.001^{*}$ \\
BruNet-LingBot & -0.0019 (-0.0219, +0.0217) & 0.5480 \\
\hline
\end{tabular}%
}
\end{minipage}

\end{table*}

As shown in Table~\ref{tab:bootstrap_ci}, BruNet-LingBot achieved the highest mean Dice and IoU, followed by BruNet-DINOv3. Table~\ref{tab:stat_tests} shows that BruNet-DINOv3 outperformed all eight non-BruNet baselines for both Dice and IoU after Holm--Bonferroni correction. This includes the strongest non-BruNet baseline, OneFormer, for which the adjusted $p$-values were 0.0044 for Dice and 0.0059 for IoU. The comparisons with SAM3 and all remaining baselines had adjusted $p$-values below 0.001.

The differences between BruNet-DINOv3 and BruNet-LingBot were not statistically significant for either Dice ($p_{\mathrm{adj}}=0.4096$) or IoU ($p_{\mathrm{adj}}=0.5480$). Their median paired differences were close to zero, and both bootstrap intervals included zero. These results indicate that the two BruNet backbone variants provide comparable image-level Dice and IoU, although BruNet-LingBot achieved the higher arithmetic mean for both metrics.

The bootstrap interval for $\Delta_{\mathrm{med}}$ and the Wilcoxon test summarise different properties of the paired differences: the former estimates the median difference, whereas the latter uses the signed ranks of all non-zero differences. Consequently, a median-difference interval can include zero even when the corrected Wilcoxon test is significant, as observed for the OneFormer comparison.

\section{Ablation Study}

\wqmnew{To evaluate the contribution of the SAM mask decoder and the intermediate feature upsampling in BruNet, we conducted two ablation experiments using the LingBot-Vision ViT-B/16 backbone. We first replaced the SAM ViT-B mask decoder with a lightweight convolutional segmentation head. We then removed the intermediate bilinear upsampling to $56\times56$ and passed the native $32\times32$ LingBot feature grid directly to the SAM decoder. All variants were trained using the same data, optimisation settings, and Dice + BCE loss, and evaluated on the same 86-image bruise dataset using the dual-region protocol.}

\begin{table*}[!tbhp]
\centering
\caption{Ablation study on the 86-image bruise dataset, evaluating the SAM mask decoder and intermediate ViT-to-SAM feature upsampling.}
\label{tab:ablation}
\begin{tabular}{lccccc}
\hline
\textbf{Model Variant} & \textbf{Precision} & \textbf{Recall} & \textbf{Dice} & \textbf{IoU} & \textbf{Accuracy} \\
\hline
LingBot + Seg. Head                & 0.866 & \textbf{0.881} & 0.848 & 0.764 & 0.950 \\
LingBot w/o intermediate upsample  & 0.939 & 0.789 & 0.839 & 0.752 & 0.947 \\
BruNet-LingBot                     & \textbf{0.946} & 0.823 & \textbf{0.867} & \textbf{0.790} & \textbf{0.956} \\
\hline
\end{tabular}
\end{table*}

\wqmnew{The lightweight segmentation head projects the patch features to 256 channels, followed by a $3\times3$ convolution to 128 channels with GroupNorm and GELU. Four additional $3\times3$ convolutional blocks and two bilinear upsampling stages are then applied before a final $1\times1$ convolution produces the segmentation mask. This replaces the SAM prompt encoder and mask decoder with a conventional convolutional decoder.}

\wqmnew{As shown in Table~\ref{tab:ablation}, replacing the SAM decoder reduced Dice from 0.867 to 0.848 and IoU from 0.790 to 0.764. The convolutional head attained the highest recall of the three variants (0.881 against 0.823), but this reflects over-segmentation rather than better
delineation, as its precision fell from 0.946 to 0.866. Removing the intermediate feature upsampling reduced Dice further to 0.839 and IoU to 0.752. Although this variant retained high precision (0.939), visual inspection showed more fragmented masks that tended to retain only high-confidence core regions, reducing recall to 0.789. This supports the role of intermediate $56\times56$ upsampling in preserving spatial continuity and recovering complete bruise regions.}

\section{Conclusion}

Bruise segmentation remains challenging due to diffuse appearance and limited annotated data. We proposed BruNet, a zero-shot transfer segmentation framework that combines a ViT-based visual encoder with a SAM mask decoder. When trained on the HAM10000 skin lesion dataset, BruNet demonstrates strong generalisation to out-of-distribution bruise images without any task-specific fine-tuning.

To the best of our knowledge, this is the first work to address bruise segmentation with the assistance of computer vision and machine learning. Our dual-region evaluation set highlights the ambiguity of bruise boundaries, and BruNet consistently outperforms CNN baselines and prompt-based SAM variants across Dice and IoU metrics. The results suggest that combining structural reasoning (via ViTs) improves robustness in visually ambiguous, under-annotated clinical targets. These findings support future research in forensic and dermatological imaging, where annotation scarcity and domain shift are major challenges.

\subsection{Future Work}

\subsubsection{Integration of foundation models}

With the rapid advancement of deep learning, foundation models such as PanDerm~\cite{Yan2024-xp} have demonstrated strong performance across a wide range of skin lesion analysis tasks. Future work could explore replacing the current ViT backbone with such domain-specific foundation models, enabling improved feature representations and more effective instruction of the SAM decoder, particularly under few-shot learning settings.



\subsubsection{More modalities}

While bruise segmentation serves as an important first step towards automated bruise analysis, clinically relevant assessment extends beyond localisation. Future research should investigate the integration of additional predictive tasks, such as estimating bruise age and severity, which could provide more comprehensive support for medical and forensic decision-making.

{
    \small
    \bibliographystyle{ieeenat_fullname}
    \bibliography{main}
}
\end{document}